\documentclass[journal]{IEEEtran}
\usepackage{amsmath}
\usepackage{amssymb}
\usepackage{latexsym}
\usepackage{psfrag}
\usepackage{graphicx,subfig}
\usepackage{epstopdf}
\usepackage{hyperref}
\usepackage{multicol}
\usepackage{fleqn}
\usepackage{enumerate}
\usepackage{color}

\usepackage{mathrsfs}

\usepackage{xcolor}
\definecolor{Orange}{RGB}{255,165,0}
\definecolor{Purple}{RGB}{144,32,144}
\newcommand{\tabref}[1]{TABLE~\ref{#1}}

\usepackage{type1cm}
\usepackage{booktabs}
\usepackage{float}
\usepackage[marginal]{footmisc}
\usepackage{array}
\usepackage{cases}

\newtheorem{remark}{Remark}
\usepackage{cite}
\allowdisplaybreaks[4]

\usepackage{algorithm}
\usepackage{algorithmic}

\begin{document}

\title{A Safety Modulator Actor-Critic Method in Model-Free Safe Reinforcement Learning and Application in UAV Hovering}
\author{Qihan Qi, Xinsong Yang, Gang Xia, Daniel W. C. Ho, \IEEEmembership{Fellow,~IEEE}, and Pengyang~Tang
\thanks{This work was supported in part by the National Natural Science Foundation of China (NSFC) under Grant Nos. 62373262 and 62303336, and in part by the Central guiding local science and technology development special project of Sichuan, and in part by the Fundamental Research Funds for Central Universities under Grant No. 2022SCU12009, and in part by the Sichuan Province Natural Science Foundation of China (NSFSC) under Grant Nos. 2022NSFSC0541,  2022NSFSC0875, 2023NSFSC1433, and in part by the National Funded Postdoctoral Researcher Program of China under Grant GZB20230467, in part by the China Postdoctoral Science Foundation under Grant 2023M742457, in part by the Power Quality Engineering Research Center of Ministry of Education under Grant KFKT202305.  Corresponding authors: Xinsong Yang.}\\
\thanks{Q.~Qi, X.~Yang, and G. Xia are with the College of Electronics and Information Engineering, Sichuan University, Chengdu 610065, China (e-mails: qiqihan@stu.scu.edu.cn (Q. Qi); xinsongyang@163.com or xinsongyang@scu.edu.cn (X. Yang); 17623110370@163.com (X. Gang)).}
\thanks{Daniel W. C. Ho is with the Department of Mathematics,
City University of Hong Kong, Kowloon, Hong Kong (e-mail:
madaniel@cityu.edu.hk).}
\thanks{P.~Tang is with the Institute of Unmanned Aerial Vehicle Systems, Lingchuan Industry, Chengdu 610100, China (e-mail:tangpy86@126.com).}
}
\maketitle
\begin{abstract}
This paper proposes a safety modulator actor-critic (SMAC) method to address safety
constraint and overestimation mitigation in model-free safe reinforcement learning (RL). A safety modulator is developed to satisfy safety constraints by modulating actions, allowing the policy to ignore safety constraint and focus on maximizing reward. Additionally, a distributional critic with a theoretical update rule for SMAC is proposed to mitigate the overestimation of Q-values with safety constraints. Both simulation and real-world scenarios experiments on Unmanned Aerial Vehicles (UAVs) hovering confirm that the SMAC can effectively maintain safety constraints and outperform mainstream baseline algorithms.
\end{abstract}


\section{Introduction}
Reinforcement learning (RL) has demonstrated remarkable achievements in games and simulations \cite{silverMasteringGameGo2016,sun2021model,vinyalsGrandmasterLevelStarCraft2019,tunyasuvunakool2020dmcontrol,
hao2023exploration,gao2024reinforcement} since results derived from it are rarely fail. However, in practical scenarios, using RL is not an easy task because various inherent risks in RL agents often deteriorate the function of RL, which may lead to  unsafe behaviors or even catastrophic consequences, such as equipment damage, environmental degradation, or even loss of human life. Therefore, it is a great challenge to guarantee the safe behavior derived by RL in real applications, especially for UAVs.

In the field of safe RL, there exist two common methods \cite{zhao2023state}: safety filter method \cite{dalal2018safe,zhao2021model} and safety learning method \cite{tessler2018reward,yang2023safety,amir2024safe,ma2024learn,wang2023safety}. The safety filter method solves the safety problem by using a safety filter on the actions of the RL agent \cite{dalal2018safe,zhao2021model}. Although the safety filter can transform unsafe actions into safe actions, it neither guarantees safety nor offers adaptability \cite{dalal2018safe} or even requires extremely precise information of a dynamic model for constructing safety filter \cite{zhao2021model}. Therefore, achieving satisfactory safety performance by using safety filters in practical systems remains challenging, and it is more difficult when different tasks require diverse modeling approaches, particularly for tasks that lack existing models. In contrast, the safety methods in learning optimizes the policy with safety constraints throughout the learning process directly. 
An advantage of safety learning methods is their model-free nature, which allows them to be applied to complex systems without requiring an accurate system model. This is particularly beneficial in real-world scenarios where accurate models are often difficult or impossible to obtain.
A notable safety in learning method is the Lagrangian method \cite{tessler2018reward,yang2023safety}, which transforms an optimization problem with safety constraints into an unconstrained primal-dual optimization problem. This is achieved by dynamically adjusting the weight of the safety cost rewards based on the degree of satisfaction with safety constraints. However, under such an approach, the policy may face a substantial burden or even fail to achieve its safety learning objectives because it needs to trade off the reward against the cost rewards. Hence, it is urgent to develop new techniques to alleviate the burden of policy while meeting safety constraints.

On the other hand, most RL algorithms tend to learn overestimated Q-values \cite{van2016deep, lee2019bias, duan2021distributional}, resulting in a suboptimal policy formulation. It is demonstrated in \cite{van2016deep} that system noise, approximation error, or any other sources can induce an overestimation bias. To mitigate overestimation, approaches like double Q-learning and double Q-network were developed by \cite{hasselt2010double, van2016deep} to leverage a target Q-network to provide unbiased estimates. However, these methods are inherently limited to discrete action spaces. Although the authors in \cite{fujimoto2018addressing} extend double Q-learning and Q-network to continuous action spaces by using actor-critic method, the overestimation problem persists due to the high similarity between the online Q-value and the target Q-value.  While distributional critic approaches have been employed \cite{yang2023safety,du2023safe}, these methods lack theoretical analysis to derive a gradient update rule that addresses overestimation. Although \cite{duan2021distributional} effectively mitigates overestimation with a theoretically guaranteed gradient update rule, their approach fails to address safety constraints, let alone alleviate the burden of policy in meeting safety constraints. These gaps motivate us to propose a novel method to investigate the mitigation of overestimation and the alleviation of policy burden in safe RL.

This paper proposes an SMAC method to address the issues of both safety constraints and mitigate overestimation. A safety modulator is introduced to modulate the action of policy, which alleviates the burden of policy and allows the policy to concentrate on maximizing the reward while disregarding the trade-off for cost rewards. The main contributions are as follows.
\begin{enumerate}
\item[{(1)}] A model-free safety modulator is presented to modulate the action of policy, which enables the policy to neglect cost rewards and focus on maximizing rewards. Without the safety modulator, the policies in \cite{tessler2018reward,yang2023safety} may suffer failures in the learning process because they always need to trade off the maximization of rewards against cost rewards.
\item[{(2)}] To mitigate overestimation, a distributional critic for SMAC is proposed to incorporate distributional information with theoretically updated rules to mitigate overestimation under safety constraints. Different from existing papers, the overestimation mitigation approach is given by detailed theoretical analysis.
\item[{(3)}]  Both PyBullet simulations and real-world experiments for UAV hovering demonstrate that the proposed SMAC algorithm can effectively mitigate overestimation while maintaining safety constraints. Comparative experiments show the merit  that our algorithm outperforms the mainstream baseline algorithms in \cite{haarnoja2018soft,ha2020learning}.
\end{enumerate}

The rest is organized as follows. In Section II presents the safety modulator for safe RL. Section III analyzes the overestimation problem and mitigates overestimation with the distributional critic. In Section IV proposes the SMAC algorithm in detail. Section V presents the UAV hovering task's setup and results to show the SMAC's efficacy. Finally, Section VI draws the conclusion.

\section{Safety modulator}
Safe RL issue can be modeled as constrained Markov decision process (CMDP) $(X,U,r,r_{c},p)$ \cite{altman2021constrained,wachi2020safe}, where $X$ and $U$ are the continuous state space and continuous action space, respectively, $r:X\times U\rightarrow[r_{\min},r_{\max}]$ is the reward function, $r_{c}:X\times U\rightarrow[c_{\min},c_{\max}]$ is the cost reward function, $p:X\times U\times X \rightarrow[0,1]$ is the state transition function. It is assumed that state $x_{t}\in X$ at the time $t$ can be observed from the environment, the agent takes an action $u_{t}\in U$ to interact with the environment and transmit state $x_{t}$ to $x_{t+1}$. The initial state $x_{0}\sim \o$, $\o$ is the initial state distribution, $\pi(\cdot|x_{t})$ is the action policy distribution under state $x_{t}$ and action $u_{t}\sim\pi(\cdot|x_{t})$. The entire trajectory distribution under policy $\pi$ is represented as $T_{\pi}=(x_{0},u_{0},x_{1},u_{1},\cdots)$.

Consider the following safe RL optimization problem with $(x_{t},u_{t})\sim T_{\pi}$
\begin{align} \label{eq1}
\max_{\pi}  \ &\mathop{\mathbb{E}}\limits_{} [\sum\limits_{t=0}^{\infty}\gamma^{t}r(x_{t},u_{t})], \\
\text{s.t.}   \mathop{\mathbb{E}}\limits_{} &[\sum\limits_{t=0}^{\infty}\gamma^{t}r_{c}(x_{t},u_{t})] \leq C, \nonumber
\end{align}
where $r(x_{t},u_{t})$ is the reward function and $r_{c}(x_{t},u_{t})$ is the cost reward function, $C\geq0$ is the given safety constraint, $\gamma$ is the discount factor of reward and cost reward.

The safe RL optimization \eqref{eq1} is actually a  constrained optimization problem. By using the Lagrangian method \cite{tessler2018reward,yang2023safety}, the constrained optimization problem can be equivalently transformed into the following unconstrained optimization one:
\begin{align*}
\min\limits_{\lambda\geq0} \max_{\pi}  \mathop{\mathbb{E}}\limits_{} [&\sum\limits_{t=0}^{\infty}\gamma^{t}r(x_{t},s_{t})-\lambda(\sum\limits_{t=0}^{\infty}\gamma^{t}r_{c}(x_{t},u_{t})-C)],
\end{align*}
where $\lambda\geq0$ is the safety weight and can be dynamically adjusted according to the satisfaction of constraints.

For the convenience of subsequent derivations, let $Q(x_{0},u_{0})=\mathop{\mathbb{E}}\limits_{} \sum\limits_{t=0}^{\infty}\gamma^{t} r(x_{t},u_{t})$ and $Q_{c}(x_{0},u_{0})=\mathop{\mathbb{E}}\limits_{} \sum\limits_{t=0}^{\infty}\gamma^{t} r_{c}(x_{t},u_{t})$. Then above unconstrained optimization problem can be simplified as
\begin{align}
  \min\limits_{\lambda\geq0} \max_{\pi} \  \mathop{\mathbb{E}}\limits_{} [&Q(x_{0},u_{0})-\lambda (Q_{c}(x_{0},u_{0})-C)]. \label{eq2}
\end{align}

 There are two steps to solve \eqref{eq2}, the first one is optimizing policy $\pi$ for given $\lambda$, second is optimizing $\lambda$ for given $\pi$:
\begin{align}
 \max_{\pi}  \  \mathop{\mathbb{E}}\limits_{} [&Q(x_{0},u_{0})-\lambda (Q_{c}(x_{0},u_{0})-C)], \label{eq3}
\end{align}
\begin{align}
 \min_{\lambda\geq0}  \  \mathop{\mathbb{E}}\limits_{} [-\lambda (Q_{c}(x_{0},u_{0})-C)]. \label{eq4}
\end{align}

\begin{remark}
According to the contraction mapping theorem in \cite{munos2016safe}, a unique fixed point exists in a
complete metric space. By continuously applying the contraction mapping, starting from any initial state $x_{0}$ and $u_{0}\sim\pi(\cdot|x_{0})$, this unique fixed point can be reached. Consequently, policy iteration will converge to the optimal value function regardless of the initial estimates. For off-policy training, the optimization \eqref{eq3} can be represented as $ \max\limits_{\pi}  \ \mathop{\mathbb{E}}\limits_{} \ [Q(x_{t},u_{t})-\lambda (Q_{c}(x_{t},u_{t})-C)]$.
\end{remark}

In order to address \eqref{eq3} for the action $u_{t}\sim \pi(\cdot|x_{t})$, one can maximize $Q(x_{t},u_{t})$ and minimize $Q_{c}(x_{t},u_{t})$. In the training step, the policy constantly trades off the $Q(x_{t},u_{t})$ against the $Q_{c}(x_{t},u_{t})$. Consequently, it may face a significant challenge or failure in its task learning. To prevent this from happening, the safety modulator $\Delta u_{t}$ and modulation function $m(\cdot): A\rightarrow A $ are presented such that $u_{t}=m(\bar{u}_{t}, \Delta u_{t})$,
where $\bar{u}_{t}\sim\pi_{\theta_{u}}(\cdot|x_{t})$ is the risky policy that disregards the potential for unsafe situations, $\Delta u_{t}\sim\pi_{\theta_{\Delta}}(\cdot|x_{t},\bar{u}_{t})$ is the safety modulator for $\bar{u}_{t}$, $\pi_{\theta_{\bar{u}}}(\cdot|x_{t})$ and $\pi_{\theta_{\Delta}}(\cdot|x_{t},\bar{u}_{t})$ denote the policy approximated with parameters $\theta_{\bar{u}}$ and $\theta_{\Delta}$, respectively. In the following statement, the overall composed policy will be denoted as $\pi_{\theta_{u}\bullet\theta_{\Delta}}$.

\begin{figure}[!h]
\centering
\includegraphics[scale=0.35]{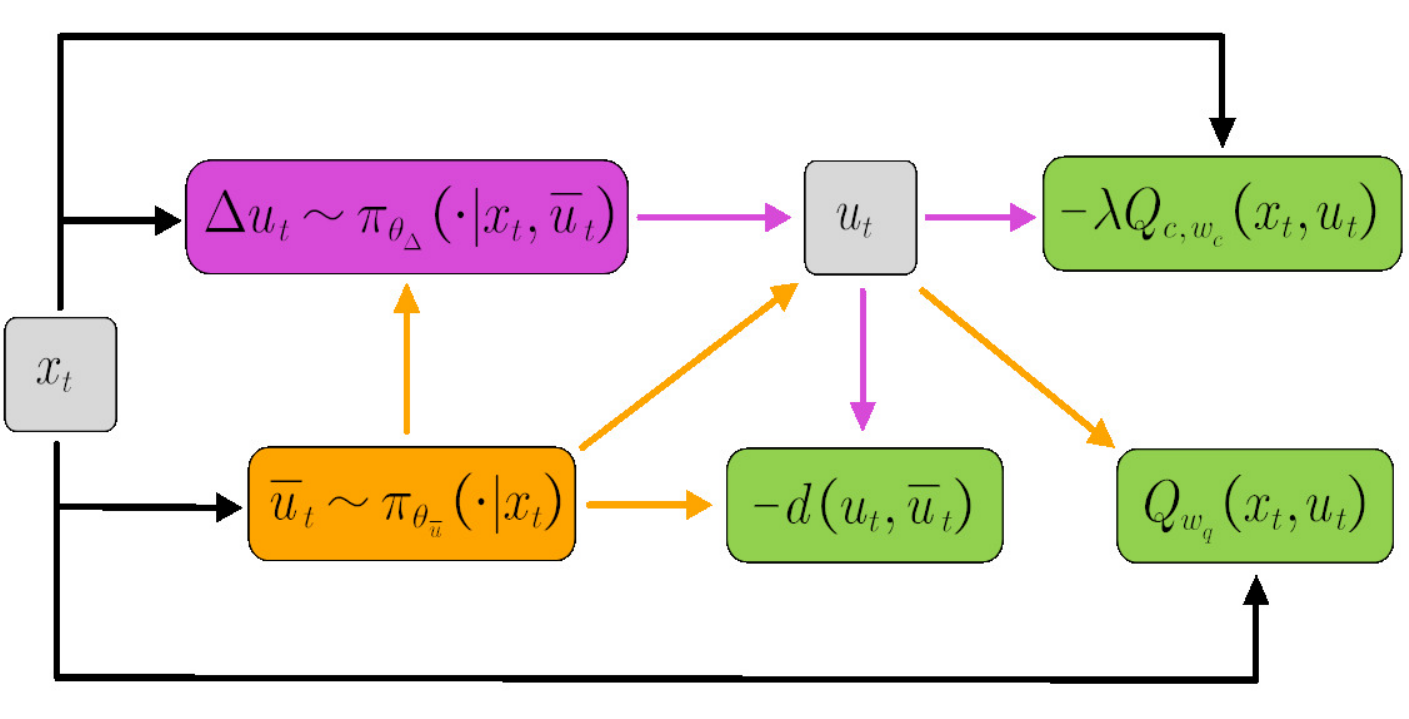}
\caption{The framework graph features nodes representing variables and edges representing operations. {\color{Orange}Orange} paths represent the gradient paths of $\theta_{\bar{u}}$, while {\color{Purple}purple} paths represent the gradient paths of $\theta_{\Delta}$. Paths depicted in black or {\color{Orange}orange} are detached for $\theta_{\Delta}$, and paths depicted in black or {\color{Purple}purple} are detached for $\theta_{\bar{u}}$.} \label{fig7}
\end{figure}

For the model training, the risky policy $\pi_{\theta_{\bar{u}}}$, safety modulator $\pi_{\theta_{\Delta}}$ and critics $Q_{w_{q}}(x_{t},u_{t})$, $Q_{c,w_{c}}(x_{t},u_{t})$ are learned from experience tuple $(x_{t},u_{t},r(x_{t},u_{t}),r_{c}(x_{t},u_{t}),x_{t+1})\sim D$, where $D$ represents the replay buffer, $Q_{w_{q}}(x_{t},u_{t})$ and $Q_{c,w_{c}}(x_{t},u_{t})$ are the approximations of $Q(x_{t},u_{t})$ and $Q_{c}(x_{t},u_{t})$ using the parameters $w_{q}$ and $w_{c}$, respectively. Introducing safety modulator, \eqref{eq3} can be divided into two parts:
\begin{align}
&(a) \max_{{\color{Orange}\theta_{u}}} \ \mathop{\mathbb{E}}\limits_{} Q_{w_{q}}(x_{t},{\color{Orange}u_{t}}),\notag\\
&(b)  \max_{{\color{Purple}\theta_{\Delta}}} \ \mathop{\mathbb{E}}\limits_{ }[ -d({\color{Purple}u_{t}},\bar{u}_{t}) -\lambda Q_{c,w_{c}}(x_{t},{\color{Purple}u_{t}})], \label{eq5}
\end{align}
where $d(u_{t},\bar{u}_{t})=\frac{1}{2}\| u_{t} - \bar{u}_{t} \|^{2}$ is the distance function between $\bar{u}_{t}$ and $u_{t}$.
The {\color{Orange}orange} ${\color{Orange}u_{t}}=m({\color{Orange}\bar{u}_{t}},{\color{Orange}\Delta u_{t}})$ is the safe action detached from the gradient of $\theta_{\Delta}$, and ${\color{Orange}\bar{u}_{t}}\sim\pi_{{\color{Orange}\theta_{\bar{u}}}}(\cdot|x_{t})$, ${\color{Orange}\Delta u_{t}} \sim \pi_{\theta_{\Delta}}(\cdot|x_{t}, {\color{Orange}\bar{u}_{t}})$.
The {\color{Purple}purple} ${\color{Purple}u_{t}}=m(\bar{u}_{t},{\color{Purple}\Delta u_{t}})$ is the safe action detached from the gradient of $\theta_{\bar{u}}$, and $\bar{u}_{t}\sim\pi_{\theta_{\bar{u}}}(\cdot|x_{t})$,
${\color{Purple}\Delta u_{t}} \sim \pi_{{\color{Purple}\theta_{\Delta}}}(\cdot|x_{t}, \bar{u}_{t})$. The framework graph is depicted in Fig. \ref{fig7}.

\begin{remark}
The modulation function $m(\bar{u}_{t},\Delta u_{t})$ is defined as $u_{t}=m(\bar{u}_{t},\Delta u_{t})=\text{clip}(\bar{u}_{t}+\Delta u_{t}, -u_{\max}, u_{\max})$, where $u_{\max}$ is the upper bound of action space, $\text{clip}(\cdot)$ is the function to constrain the values of $\bar{u}_{t}+\Delta u_{t}$ within a specified range $[-u_{\max}, u_{\max}]$. The modulation function can provide both flexibility and control in modifying actions, and hence, it is suitable for varying conditions by using an easy-to-implement additive safety modulator while ensuring that modifications remain within safe and acceptable boundaries.
\end{remark}

\begin{remark}
To introduce the safety modulator that allows the policy to concentrate on maximizing the reward, it is necessary to transform \eqref{eq3} into \eqref{eq5}. However, the policies in \cite{tessler2018reward,yang2023safety} fail to derive \eqref{eq5} as they cannot establish the connection between $\bar{u}$ and $\Delta u$. This paper addresses this issue by constructing the distance $d(u_{t}, \bar{u}{t}) = \frac{1}{2}| u{t} - \bar{u}_{t} |^{2}$, which enables the safety modulator to minimize this distance, thereby adjusting $\bar{u}$ as minimally as possible while still ensuring it meets certain constraints to guarantee the action's safety.
\end{remark}

The safety weight $\lambda$ in \eqref{eq4} can be optimized by minimizing the following loss $J(\lambda)$ with the entire $\mathbf{M}$ steps episode state-action pairs $\{(s_{i},a_{i})\}_{i=0}^{\mathbf{M}-1}\sim T_{\pi_{\theta_{u}\bullet\theta_{\Delta}}}$,
\begin{align}
    J_{\lambda}(\lambda) = \lambda(C- \sum\limits_{t=0}^{\mathbf{M}-1}\gamma^{t} r_{c}(x_{t},u_{t})).
\end{align}

After each rollout, we collect a batch of cost rewards to guarantee that the safety constraint is strictly satisfied. The $\lambda$ is only updated after collecting the entire episode state-action pairs.

\section{Overestimation mitigation}
In this section, the issue of overestimation inherent in Q-learning is discussed, and specific overestimation value is provided through formula derivation. After that, the distributional critic and corresponding update rule are introduced to mitigate overestimation.

The Q-value approximated by the parameter $w_{q}$ is expressed as $Q_{w_{q}}(x_{t},u_{t})=\overline{Q}(x_{t},u_{t})+\nu_{t}$ with $\nu_{t}$ being a random zero mean noise, $\overline{Q}(x_{t},u_{t})$ is the ideal Q-value without bias. Then, the updated parameter $w_{q}'$ can be obtained by the following formula
\begin{align*}
w_{q}'=w_{q}+\eta(y-Q_{w_{q}}(x_{t},u_{t}))\nabla_{w_{q}}Q_{w_{q}}(x_{t},u_{t}),
\end{align*}
where $\eta$ is the learning rate which controls update step size, $y=\mathop{\mathbb{E}}\limits_{}[r(x_{t},u_{t}) + \gamma \mathop{\max}\limits_{u_{t+1}}Q_{w_{q}}(x_{t+1},u_{t+1})]$ is the Bellman equation.

Similarly, the updated parameter $\overline{w}'_{q}$ of the $w'_{q}$ is formulated as
\begin{align*}
\overline{w}'_{q}=w_{q}+\eta(\overline{y}-Q_{w_{q}}(x_{t},u_{t}))\nabla_{w_{q}}Q_{w_{q}}(x_{t},u_{t}),
\end{align*}
where $\overline{y}=\mathop{\mathbb{E}}\limits_{} \ [r(x_{t},u_{t}) + \gamma \mathop{\max}\limits_{u_{t+1}}\overline{Q}(x_{t+1},u_{t+1})]$ is the ideal value of $y$.

Employing first-order  Taylor's expansion, the updated values of $Q_{w_{q}}$ and $Q_{\overline{w}_{q}}$ can be approximated by the following $Q_{w_{q}'}(x_{t}, u_{t})$ and $Q_{\overline{w}_{q}'}(x_{t}, u_{t})$, respectively.
\begin{align}
Q_{w_{q}'}(x_{t}, u_{t}) \approx &Q_{w_{q}}(x_{t}, u_{t}) + \eta (y - Q_{w_{q}}(x_{t}, u_{t}))\notag\\
 &\lVert \nabla_{w_{q}} Q_{w_{q}}(x_{t}, u_{t}) \rVert^2, \notag \\
Q_{\overline{w}_{q}'}(x_{t}, u_{t}) \approx &Q_{w_{q}}(x_{t}, u_{t}) + \eta (\overline{y} - Q_{w_{q}}(x_{t}, u_{t}))\notag\\ &\lVert \nabla_{w_{q}} Q_{w_{q}}(x_{t}, u_{t}) \rVert^2.
\label{eq3.3}
\end{align}

Then, the estimation error of $Q_{w_{q}}$ during an update step is
\begin{align*}
\varepsilon(x_{t},u_{t})=&\mathop{\mathbb{E}}\limits_{ } [Q_{w_{q}'}(x_{t}, u_{t})-Q_{\overline{w}_{q}'}(x_{t}, u_{t})] \\
\approx&\mathop{\mathbb{E}}\limits_{} [\eta(y-\overline{y})\lVert \nabla_{w_{q}} Q_{w_{q}}(x_{t}, u_{t}) \rVert^2] \\
=&\eta\gamma\mathop{\mathbb{E}}\limits_{} [\mathop{\max}\limits_{u_{t+1}}Q_{w_{q}}(x_{t+1},u_{t+1}) -\mathop{\max}\limits_{u_{t+1}}\overline{Q}(x_{t+1},u_{t+1})]\notag\\
&\times\lVert \nabla_{w_{q}} Q_{w_{q}}(x_{t}, u_{t}) \rVert^2.
\end{align*}

Considering $Q_{w_{q}}(x_{t},u_{t})=\overline{Q}(x_{t},u_{t})+\nu_{t}$ and letting $\epsilon=\mathop{\mathbb{E}}\limits_{ } [\mathop{\max}\limits_{u_{t+1}}[\overline{Q}(x_{t+1},u_{t+1})+\nu_{t+1}]
-\mathop{\max}\limits_{u_{t+1}}\overline{Q}(x_{t+1},u_{t+1})]$, one has
\begin{align*}
\varepsilon(x_{t},u_{t})\approx\eta\gamma\epsilon\lVert \nabla_{w_{q}} Q_{w_{q}}(x_{t}, u_{t}) \rVert^2.
\end{align*}

It is noteworthy that $\epsilon\geq0$ \cite{hasselt2010double,thrun2014issues}, which implies  $\varepsilon(x_{t},u_{t})\geq0$, \emph{i.e.}, the max operator inherently introduces an upward bias to estimation errors. Even if a single update introduces only a slight upward bias, the cumulative effect of these bias through temporal difference (TD) learning can lead to substantial overestimation, which makes the policy suboptimal.

To mitigate overestimation, a distributional critic denoted by $\mathcal{Z}(x_{t},u_{t})$ is considered, which follows a normal distribution $Z(\cdot|x_{t},u_{t})$. The mean and standard deviation of this distribution are approximated by the neural network outputs $Q_{w_{q}}(x_{t},u_{t})$ and $\sigma_{w_{\sigma}}(x_{t},u_{t})$, respectively. Define $Z(\cdot|x_{t},u_{t})= N(Q_{w_{q}}(x_{t},u_{t}), \sigma_{w_{\sigma}}^{2}(x_{t},u_{t}))$.

Consider the distributional Bellman equation $\widetilde{y} = r + \gamma \mathcal{Z}(x_{t+1}, \overline{u}_{t+1})$, where $\overline{u}_{t+1}=\arg\mathop{\max}\limits_{u_{t+1}}  $ \\ $ Q_{w_{q}}(x_{t+1},u_{t+1})$.  Assuming $\widetilde{y}\sim \overline{Z}(\cdot|x_{t},u_{t})$ with $\overline{Z}(\cdot|x_{t},u_{t})$ being the ideal normal distribution, one has
\begin{align*}
\mathbb{E} [\widetilde{y}]= \mathop{\mathbb{E}}\limits_{} [r(x_{t},u_{t}) + \gamma \mathop{\max}\limits_{u_{t+1}}Q_{w_{q}}(x_{t+1},u_{t+1})]=y.
\end{align*}

For convenience of later study, let $\overline{Z}(\cdot | x_t, u_t) = N(y, \overline{\sigma}^2)$, where $\overline{\sigma}$ represents the ideal standard deviation.

To measure the distance between $\overline{Z}(\cdot|x_{t},u_{t})$ and $Z(\cdot|x_{t},u_{t})$, the Kullback-Leibler (KL) divergence \cite{duan2021distributional,bellemare2017distributional,barth2018distributed} is utilized. Since both the distributions are normal, the KL divergence can be analytically expressed as follows
\begin{align}\label{eq4_KL}
&D_{KL}(\overline{Z}(\cdot|x_{t},u_{t}), Z(\cdot|x_{t},u_{t})) = \log \frac{\sigma_{w_{\sigma}}(x_{t},u_{t})}{\overline{\sigma}}\notag\\
 &+\frac{\overline{\sigma}^{2}(x_{t},u_{t})+(y-Q_{w_{q}}(x_{t},u_{t}))^{2}}{2\sigma_{w_{\sigma}}^{2}(x_{t},u_{t})}-\frac{1}{2}.
\end{align}

As a result, the parameters $w_{q}$ and $w_{\sigma}$ are updated as follows
\begin{align}
w_{q}' &= w_{q} + \eta \nabla_{w_{q}} D_{KL}(\overline{Z}(\cdot|x_{t},u_{t}), Z(\cdot|x_{t},u_{t})), \nonumber  \\
&= w_{q} + \eta \frac{\widetilde{y} - Q_{w_{q}}(x_{t},u_{t})}{\sigma_{w_{\sigma}}(x_{t},u_{t})^2} \nabla_{w_{q}} Q_{w_{q}}(x_{t},u_{t}), \label{eq4_update1} \\
w_{\sigma}' &= w_{\sigma} + \eta \nabla_{w_{\sigma}} D_{KL}(\overline{Z}(\cdot|x_{t},u_{t}), Z(\cdot|x_{t},u_{t})), \nonumber  \\
&= w_{\sigma} + \eta \frac{\overline{\sigma}^2 - \sigma_{w_{\sigma}}^{2}(x_{t},u_{t}) + (y - Q_{w_{q}}(x_{t},u_{t}))^2}{\sigma_{w_{\sigma}}(x_{t},u_{t})^3} \notag\\
&\times\nabla_{w_{\sigma}} \sigma_{w_{\sigma}}(x_{t},u_{t}). \label{eq4_update2}
\end{align}

Additionally, there exists an ideal target $\widetilde{y}$, denoted as $\overline{\widetilde{y}}$, such that $\mathbb{E} [\overline{\widetilde{y}}]=\mathop{\mathbb{E}}\limits_{} [r(x_{t},u_{t}) + \gamma \mathop{\max}\limits_{u_{t+1}}\overline{Q}(x_{t+1},u_{t+1})]=\overline{y}$. Following a similar derivation to the KL divergence \eqref{eq4_KL}, the update for $\overline{w}_{q}'$ is given by
\begin{align}\label{eq4_update3}
\overline{w}_{q}'=w_{q}+\eta \frac{\overline{y} - Q_{w_{q}}(x_{t},u_{t})}{\sigma_{w_{\sigma}}(x_{t},u_{t})^2} \nabla_{w_{q}} Q_{w_{q}}(x_{t},u_{t}).
\end{align}

In a manner similar to the derivation of $\varepsilon(x_{t},u_{t})$, the overestimation bias of $Q_{w_{q}}(x_{t},u_{t})$ in the distributional critic $\mathcal{Z}(x_{t},u_{t})$ can be expressed as
\begin{align}\label{eq4_bias}
\widetilde{\varepsilon}(x_{t},u_{t})= \frac{\varepsilon(x_{t},u_{t})}{\sigma_{w_{\sigma}}^{2}(x_{t},u_{t})}.
\end{align}

\begin{remark}
According to \eqref{eq4_bias}, the overestimation bias $\widetilde{\varepsilon}(x_{t},u_{t})$ is inversely proportional to $\sigma_{w_{\sigma}}^{2}(x_{t},u_{t})$. It is obvious that once $\sigma_{w_{\sigma}}(x_{t},u_{t})\geq1$, the condition $\widetilde{\varepsilon}(x_{t},u_{t})\leq\varepsilon(x_{t},u_{t})$ can be guaranteed, and hence the overestimation can be mitigated. Therefore, we choose $\sigma_{w_{\sigma}}(x_{t},u_{t})=\max(\sigma_{w_{\sigma}}(x_{t},u_{t}), \sigma_{\min})$, where $\sigma_{\min}\geq1$ is a given parameter.

\end{remark}

\begin{remark}
It should be noted that the safety constraint $C$ is a given deterministic constant. Intuitively, using a distributional cost critic to evaluate the deterministic $Q_{c}(x_{t},u_{t})$ is unsuitable. Therefore, we only use a distributional critic for $Q(x_{t},u_{t})$.
\end{remark}

\section{Safety modulator actor-critic}
This section proposes an SMAC algorithm, incorporating the corresponding update rules for the risky policy $\pi_{\theta_{\bar{u}}}(\cdot|x_{t})$, the safety modulator $\pi_{\theta_{\Delta}}(\cdot|x_{t},\bar{u}{t})$, the distributional critic $Z{w_{q}}(\cdot|x_{t},u_{t})$, and the cost critic $Q_{c,w_{c}}(\cdot|x_{t},u_{t})$, with approximate parameters $\theta_{\bar{u}}$, $\theta_{\Delta}$, $w_{q}$, and $w_{c}$. It is noteworthy that the update rule of the distributional critic in \textit{Distributional Policy Evaluation} can theoretically guarantee overestimation mitigation. Additionally, a series of training techniques are employed in \textit{Distributional Policy Evaluation} to improve training stability. The updated rule of the safety modulator is detached from the gradient  $\theta_{\bar{u}}$ to alleviate the burden of risky policy to focus on maximizing rewards.

\subsection{Safety policy evaluation}
\subsubsection{Distributional policy evaluation}

Considering $\mathcal{B}_{\pi_{\tilde{\theta}_{\bar{u}}\bullet\tilde{\theta}_{\Delta}}}\mathcal{Z}(x_{t},u_{t}) \sim \mathcal{B}_{\pi_{\tilde{\theta}_{\bar{u}}\bullet\tilde{\theta}_{\Delta}}}Z_{\widetilde{w}_{q}}(\cdot|x_{t},u_{t})$, $(x_{t},u_{t})\sim D$, the loss function of KL divergence is given as
\begin{align}\label{eq5_spe}
J_{z}&(w_{q}) = \mathop{\mathbb{E}}\limits_{} [D_{KL}(\mathcal{B}_{\pi_{\tilde{\theta}_{\bar{u}}\bullet\tilde{\theta}_{\Delta}}}Z_{\widetilde{w}_{q}}(\cdot|x_{t},u_{t}),Z_{w_{q}}(\cdot|x_{t},u_{t}))] \nonumber \\
=& \mathop{\mathbb{E}}\limits_{} \bigg[\int[\log(P(\mathcal{B}_{\pi_{\tilde{\theta}_{\bar{u}}\bullet\tilde{\theta}_{\Delta}}}\mathcal{Z} (x_{t},u_{t})| \mathcal{B}_{\pi_{\tilde{\theta}_{\bar{u}}\bullet\tilde{\theta}_{\Delta}}} Z_{\widetilde{w}_{q}}(\cdot|x_{t},u_{t}))) \nonumber \\
&- \log(P(\mathcal{B}_{\pi_{\tilde{\theta}_{\bar{u}}\bullet\tilde{\theta}_{\Delta}}}\mathcal{Z} (x_{t},u_{t}) | Z_{w_{q}}(\cdot|x_{t},u_{t})) )] \notag\\
&P(\mathcal{B}_{\pi_{\tilde{\theta}_{\bar{u}}\bullet\tilde{\theta}_{\Delta}}}\mathcal{Z} (x_{t},u_{t}) |\mathcal{B}_{\pi_{\tilde{\theta}_{\bar{u}}\bullet\tilde{\theta}_{\Delta}}}Z_{\widetilde{w}_{q}}(\cdot|x_{t},u_{t}))\bigg]\notag\\
&d\mathcal{B}_{\pi_{\tilde{\theta}_{\bar{u}}\bullet\tilde{\theta}_{\Delta}}}\mathcal{Z} (x_{t},u_{t})
\nonumber \\
=& \mathop{\mathbb{E}}\limits_{} [- \log(P(\mathcal{B}_{\pi_{\tilde{\theta}_{\bar{u}}\bullet\tilde{\theta}_{\Delta}}}\mathcal{Z} (x_{t},u_{t}) | Z_{w_{q}}(\cdot|x_{t},u_{t})) )]+\Im,
\end{align}
where $\Im$ is independent of the optimized parameter $w_{q}$, $\widetilde{w}_{q}$ is the parameter of target distribution $Z_{\widetilde{w}_{q}}(\cdot|x_{t},u_{t})$, $\mathcal{B}_{\pi_{\tilde{\theta}_{\bar{u}}\bullet\tilde{\theta}_{\Delta}}}$ is the Bellman operator with policy $\pi_{\tilde{\theta}_{\bar{u}}\bullet\tilde{\theta}_{\Delta}}$, and $\pi_{\tilde{\theta}_{\bar{u}}\bullet\tilde{\theta}_{\Delta}}$ is the safe target policy with target parameters $\tilde{\theta}_{\bar{u}}$ and $\tilde{\theta}_{\Delta}$.

In view of $Z_{w_{q}}(\cdot|x_{t},u_{t})=N(Q_{w_{q}}(x_{t},u_{t}), \sigma_{w_{q}}^{2}(x_{t},u_{t}))$, the gradient of $J_{z}(w_{q})$ is obtained as
\begin{align}\label{eq5_gdwqr}
\nabla_{w_{q}}& J_{z}(w_{q}) = \mathop{\mathbb{E}} [- \nabla_{w_{q}} \log(P(\mathcal{B}_{\pi_{\tilde{\theta}_{\bar{u}}\bullet\tilde{\theta}_{\Delta}}}\mathcal{Z} (x_{t},u_{t}) | Z_{w_{q}}(\cdot|x_{t},u_{t})) )]  \nonumber \\
=& \mathop{\mathbb{E}} \bigg[- \nabla_{w_{q}} \log\bigg(\frac{\exp(-\frac{(\mathcal{B}_{\pi_{\tilde{\theta}_{\bar{u}}\bullet\tilde{\theta}_{\Delta}}}\mathcal{Z}(x_{t},u_{t})-Q_{w_{q}}(x_{t},u_{t}))^{2}} {2\sigma_{w_{q}}^{2}(x_{t},u_{t})})}{\sqrt{2\pi}\sigma_{w_{q}}(x_{t},u_{t})}\bigg)\bigg] \nonumber \\
=&\mathop{\mathbb{E}} \bigg[- \nabla_{w_{q}} \bigg( \frac{(\mathcal{B}_{\pi_{\tilde{\theta}_{\bar{u}}\bullet\tilde{\theta}_{\Delta}}}\mathcal{Z}(x_{t},u_{t} )-Q_{w_{q}}(x_{t},u_{t}))^{2}}{2\sigma_{w_{q}}^{2}(x_{t},u_{t})} \notag\\
&+ \log \sigma_{w_{q}}(x_{t},u_{t}) + \log\sqrt{2\pi}
\bigg)  \bigg] \notag \\
=&\mathop{\mathbb{E}} \bigg[-\frac{\widetilde{y}-Q_{w_{q}}(x_{t},u_{t})}{2\sigma_{w_{q}}^{2}(x_{t},u_{t})}\nabla_{w_{q}}Q_{w_{q}}(x_{t},u_{t})
 \nonumber \\
&-\frac{ - \sigma_{w_{q}}^{2}(x_{t},u_{t}) + (\widetilde{y}-Q_{w_{q}}(x_{t},u_{t}))^{2}}{\sigma_{w_{q}}^{3}(x_{t},u_{t})} \notag\\ &\times\nabla_{w_{q}} \sigma_{w_{q}}(x_{t},u_{t})\bigg].
\end{align}

Inspired by \cite{fujimoto2018addressing,duan2021distributional,duan2023dsac}, the independent double Q-networks for critic are used, which are $Q_{w_{q}^{1}}$ and $Q_{w_{q}^{2}}$. The critic tends to choose the smaller mean value between $Q_{w_{q}^{1}}$ and $Q_{w_{q}^{2}}$. Meanwhile, the clip function is used in $(\widetilde{y}-Q_{w_{q}}(x_{t},u_{t}))^{2}$ to avoid gradient explosion. Moreover, the mean value of $\widetilde{y}$ keeps training stable since  $\mathcal{Z}(x_{t+1},u_{t+1})$ is sampled from distribution $Z(\cdot|x_{t+1},u_{t+1})$. With the help of these steps, the corresponding update rule of stable gradient $\nabla_{w_{q}^{i}} J_{z}(w_{q}^{i})$, $i=1,2$ can be represented as follows
\begin{align}\label{eq5_gdt}
\nabla_{w_{q}^{i}} J_{z}&(w_{q}^{i})\approx\mathop{\mathbb{E}}\limits_{ } \bigg[-\frac{\hat{y}_{w_{q}}^{\min}-Q_{w_{q}^{i}}(x_{t},u_{t})}{2\sigma_{w_{q}^{i}}^{2}(x_{t},u_{t})}\nabla_{w_{q}^{i}}Q_{w_{q}^{i}}(x_{t},u_{t})
 \nonumber \\ &-\frac{ - \sigma_{w_{q}^{i}}^{2}(x_{t},u_{t}) + (\Delta_{w_{q}^{i}})^2}{\sigma_{w_{q}^{i}}^{3}(x_{t},u_{t})} \nabla_{w_{q}^{i}} \sigma_{w_{q}^{i}}(x_{t},u_{t})\bigg],
\end{align}
where
$\hat{y}_{w_{q}}^{\min}=r(x_{t},u_{t})+\gamma \mathop{\min}\limits_{i=1,2} Q_{w_{q}^{i}}(x_{t+1},u_{t+1})$,
$\Delta_{w_{q}^{i}}=\text{clip}(\widetilde{y}_{w_{q}}^{\min}- Q_{w_{q}^{i}}(x_{t},u_{t}), -\zeta \hat{\sigma}_{w_{q}^{i}}(x_{t},u_{t}), \zeta \hat{\sigma}_{w_{q}^{i}}(x_{t},u_{t}))$, $\widetilde{y}_{w_{q}}^{\min}=r(x_{t},u_{t})+\gamma \mathop{\min}\limits_{i=1,2} \mathcal{Z}_{w_{q}^{i}}(x_{t+1},u_{t+1})$,
$\hat{\sigma}_{w_{q}^{i}}(x_{t},u_{t})=\mathop{\mathbb{E}}\limits_{ } \ [\sigma_{w_{q}^{i}}(x_{t},u_{t})]$, $\zeta$ is an adjustable constant to make sure that $|\widetilde{y}_{w_{q}^{i}}-Q_{w_{q}^{i}}(x_{t},u_{t})|\leq\zeta\hat{\sigma}_{w_{q}^{i}}(x_{t},u_{t})$. Specifically, $\zeta=3$ denotes that 3-sigma rule in normal distribution.

\begin{remark}
Although the works in \cite{yang2023safety,bellemare2017distributional,barth2018distributed} employ distributional critics, they lack the update rule derived in this paper, rendering them unable to theoretically guarantee the mitigation of overestimation. Moreover, the distributional critic utilized in this paper is general, enabling the approximation of the critic with a normal distribution, even if the critic does not follow a normal distribution.
\end{remark}

\subsubsection{Cost evaluation}
Given double cost return $Q_{c,w_{c}^{i}}(x_{t},u_{t})$, $i=1,2$, define loss function $J_{c}(w_{c}^{i})$ as
\begin{align*}
J_{c}(w_{c}^{i})=&\mathop{\mathbb{E}}\limits_{ } [0.5(r_{c}(x_{t},u_{t})+\gamma \mathop{\max}\limits_{u_{t+1}\sim\pi_{\tilde{\theta}_{\bar{u}}\bullet\tilde{\theta}_{\Delta}}} Q_{c,\tilde{w}_{c}}(x_{t+1},u_{t+1}) \notag\\
&- Q_{c,w_{c}^{i}}(x_{t},u_{t}))^{2}],
\end{align*}
where $\tilde{w}_{c}$ represents the target parameter. The corresponding gradient is given by
\begin{align*}
\nabla_{w_{c}^{i}}J_{c}(w_{c}^{i})=&\mathop{\mathbb{E}}\limits_{ } [(Q_{c,w_{c}^{i}}(x_{t},u_{t})-r_{c}(x_{t},u_{t})\\
&-\gamma Q_{c,\tilde{w}_{c}}(x_{t},u_{t}))\nabla_{w_{c}}Q_{c,w_{c}}(x_{t},u_{t})].
\end{align*}

\begin{algorithm}[!h]
	\caption{SMAC Algorithm}
    \label{alg1}
	\begin{algorithmic}[1]
        \REQUIRE Initialized network parameters $\theta_{\bar{u}}$, $\tilde{\theta}_{\bar{u}}$, $\theta_{\Delta}$, $\tilde{\theta}_{\Delta}$, $w_{q}^{i}$, $\tilde{w}_{q}$, $w_{c}^{i}$, $\tilde{w}_{c}$, $i=1,2$, target update rate $\tau$, learning rate $\eta_{\bar{u}}$, $\eta_{\Delta u}$, $\eta_{q}$, $\eta_{c}$, $\eta_{\lambda}$, total training steps $\mathbf{M}$, safety weight update frequency $k$.
        \ENSURE Safe policy $\pi_{\theta_{\bar{u}}\bullet\theta_{\Delta u}}$.
		\STATE Set current training step $\mathbf{m}=0$
        \WHILE {$\mathbf{m}<\mathbf{M}$}
        \STATE Observe state $x_{t}$
        \STATE Select action $\bar{u}_{t}\sim\pi_{\theta_{\bar{u}}}(\cdot|x_{t})$ and safe modulation action $\Delta u_{t}\sim\pi_{\theta_{\Delta u}}(\cdot|x_{t},\bar{u}_{t})$
        \STATE Calculate $u_{t}=m(\bar{u}_{t},\Delta u_{t})$
        \STATE Observe reward $r(x_{t},u_{t})$, cost reward $r_{c}(x_{t},u_{t})$ and next state $x_{t+1}$
        \STATE Store tuple $(x_{t},u_{t},r(x_{t},u_{t}),r_{c}(x_{t},u_{t}),x_{t+1})$ in Replay Buffer $D$
        \IF{rollout and $(\mathbf{m}\mod k)==0$}
        \STATE Update safety weight $\lambda\leftarrow\lambda-\eta_{\lambda}\nabla_{\lambda}J_{\lambda}(\lambda)$
        \ENDIF
        \STATE Sample batch tuples $(x_{t},u_{t},r(x_{t},u_{t}),r_{c}(x_{t},u_{t}),x_{t+1})$ from $D$
        \STATE Update distributional critic $w_{q}^{i}\leftarrow w_{q}^{i}-\eta_{q}\nabla_{w_{q}^{i}} J_{z}(w_{q}^{i})$, $i=1,2$
        \STATE Update cost critic $w_{c}^{i}\leftarrow w_{c}^{i}-\eta_{c}\nabla_{w_{c}^{i}} J_{c}(w_{c}^{i})$, $i=1,2$
        \STATE Update risky policy $\theta_{\bar{u}}\leftarrow\theta_{\bar{u}}+\eta_{\bar{u}}\nabla_{\theta_{\bar{u}}}J_{\pi_{\bar{u}}}(\theta_{\bar{u}})$
        \STATE Update safety modulator $\theta_{\Delta u}\leftarrow\theta_{\Delta u}+\eta_{\Delta u}\nabla_{\theta_{\bar{u}}}J_{\pi_{\Delta u}}(\theta_{\Delta u})$
        \STATE Update target networks: \\
         $\tilde{w}_{q}\leftarrow(1-\tau)\tilde{w}_{q}+\tau w_{q}$,
         $\tilde{w}_{c}\leftarrow(1-\tau)\tilde{w}_{c}+\tau w_{c}$, \\
         $\tilde{\theta}_{\bar{u}}\leftarrow(1-\tau)\tilde{\theta}_{\bar{u}}+\tau\theta_{\bar{u}}$,
        $\tilde{\theta}_{\Delta u}\leftarrow(1-\tau)\tilde{\theta}_{\Delta u}+\tau\theta_{\Delta u}$
        \STATE$\mathbf{m}=\mathbf{m}+1$
        \ENDWHILE
	\end{algorithmic}
\end{algorithm}

\subsection{Policy improvement}
\subsubsection{Distributional risky policy improvement}
Since {\color{Orange}orange} ${\color{Orange}u_{t}}=m({\color{Orange}\bar{u}_{t}},{\color{Orange}\Delta u_{t}})$ is the safe action detached the gradient of $\theta_{\Delta}$, and ${\color{Orange}\bar{u}_{t}}\sim\pi_{{\color{Orange}\theta_{\bar{u}}}}(\cdot|x_{t})$, ${\color{Orange}\Delta u_{t}} \sim \pi_{\theta_{\Delta}}(\cdot|x_{t}, {\color{Orange}\bar{u}_{t}})$.
The risky policy can be improved by maximizing the following distributional objective
\begin{align*}
J_{\pi_{\bar{u}}}(\theta_{\bar{u}})
=\mathop{\mathbb{E}}\limits_{} [Q_{w_{q}}(x_{t},{\color{Orange}u_{t}})].
\end{align*}

It should be noted that the action ${\color{Orange}\bar{u}_{t}}$ is sampled from a Gaussian distribution, which is non-differentiable. Thus, to address this, the reparameterization trick is employed. This technique involves sampling from a standard normal distribution and scaling the sample by the standard deviation and adding the mean, which can be represented as $\bar{u}_{t}=f_{\theta_{\bar{u}}}(\varsigma_{\bar{u}_{t}};x_{t})=\bar{u}_{t,\text{mean}}+\varsigma_{\bar{u}_{t}}\odot\bar{u}_{t,\text{std}}$, where $\varsigma_{\bar{u}_{t}}$ follows a standard normal distribution, $\odot$ is the Hadamard product, $\bar{u}_{t,\text{mean}}$ and $\bar{u}_{t,\text{std}}$ are the mean and standard deviation of policy $\pi_{\theta_{\bar{u}}}(\cdot|x_{t})$, respectively. Consequently, the corresponding gradient is given by
\begin{align*}
\nabla_{\theta_{\bar{u}}}J_{\pi_{\bar{u}}}(\theta_{\bar{u}})=\mathop{\mathbb{E}}\limits_{ } [\nabla_{\theta_{\bar{u}}}
f_{\theta_{\bar{u}}}(\varsigma_{t};x_{t})\nabla_{\bar{u}_{t}}Q_{w_{q}}(x_{t},{\color{Orange}u_{t}})].
\end{align*}

\subsubsection{Safe modulator policy improvement}
Since the {\color{Purple}purple} ${\color{Purple}u_{t}}=m(\bar{u}_{t},{\color{Purple}\Delta u_{t}})$ is the safe action detached the gradient of $\theta_{\bar{u}}$, and $\bar{u}_{t}\sim\pi_{\theta_{\bar{u}}}(\cdot|x_{t})$,
${\color{Purple}\Delta u_{t}} \sim \pi_{{\color{Purple}\theta_{\Delta}}}(\cdot|x_{t}, \bar{u}_{t})$.
The safety modulator can be improved by maximizing the following objective with the given $\lambda$
\begin{align*}
J_{\pi_{\Delta u}}(\theta_{\Delta})=\mathop{\mathbb{E}}\limits_{} [-d({\color{Purple}u_{t}},\bar{u}_{t}) -\lambda Q_{c,w_{c}}(x_{t},{\color{Purple}u_{t}})].
\end{align*}

Similar to the reparameterization trick in distributional risky policy improvement and $\Delta u_{t}=f_{\theta_{\Delta}}(\varsigma_{\bar{u}_{t}};x_{t})$, one has
\begin{align*}
\nabla_{\theta_{\Delta}}J_{\pi_{\Delta u}}(\theta_{\Delta})=\mathop{\mathbb{E}}\limits_{} [-\nabla_{\theta_{\Delta}}d({\color{Purple}u_{t}},\bar{u}_{t}) - \lambda \nabla_{\theta_{\Delta}}Q_{c,w_{c}}(x_{t},{\color{Purple}u_{t}})].
\end{align*}

The detailed SMAC algorithm for alleviating risky policy and mitigating overestimation is presented in Algorithm \ref{alg1}.

\section{Experiments} \

In this section, Crazyflie 2.1 is utilized to carry out the UAV hovering experiments, where both numerical simulation and real-world experiment verify the effectiveness and safety of the SMAC.

\subsection{Simulation setup}
\begin{figure}[!h]
\centering
\includegraphics[scale=0.4]{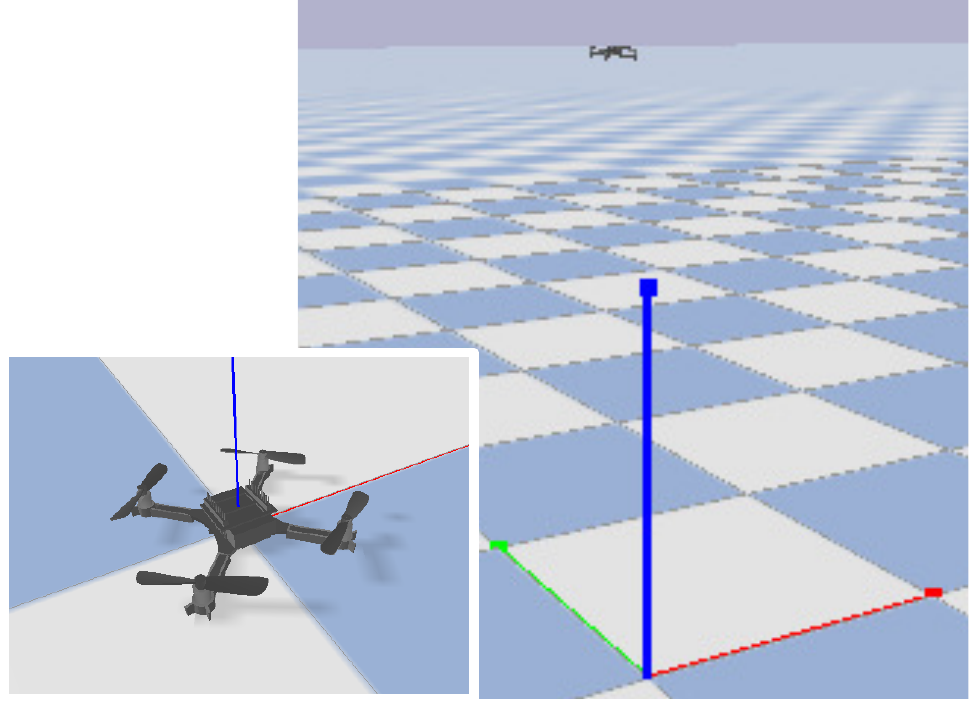}
\caption{The Crazyflie 2.1 in PyBullet.} \label{fig1}
\end{figure}

For the simulation part, the training environment is provided by PyBullet, as shown in Fig. \ref{fig1}.
Model is obtained from the URDF file; detailed information is presented in TABLE \tabref{tab1}. Notably, the simulation parameters in the URDF file are measured from real-world measurements. This makes our simulation results convenient for sim-to-real transfer.

\begin{table*}
  \centering
  \caption{Crazyflie 2.1 Parameters} \label{tab1}
\begin{tabular}{>{\centering\arraybackslash}m{0.2\textwidth} >{\centering\arraybackslash}m{0.2\textwidth}}
\toprule
Parameters &  Values  \\
\hline
Mass  & 28 g  \\
Arm  &  3.97 cm \\
Propeller radius & 2.31 cm \\
Max speed & 30 km/h \\
Thrust2weight & 1.88 \\
Hovering position & $(0 \text{m},0 \text{m},1.5 \text{m})^{T}$ \\
\bottomrule
\end{tabular}
\end{table*}

\begin{figure}[!h]
\centering
\includegraphics[scale=0.35]{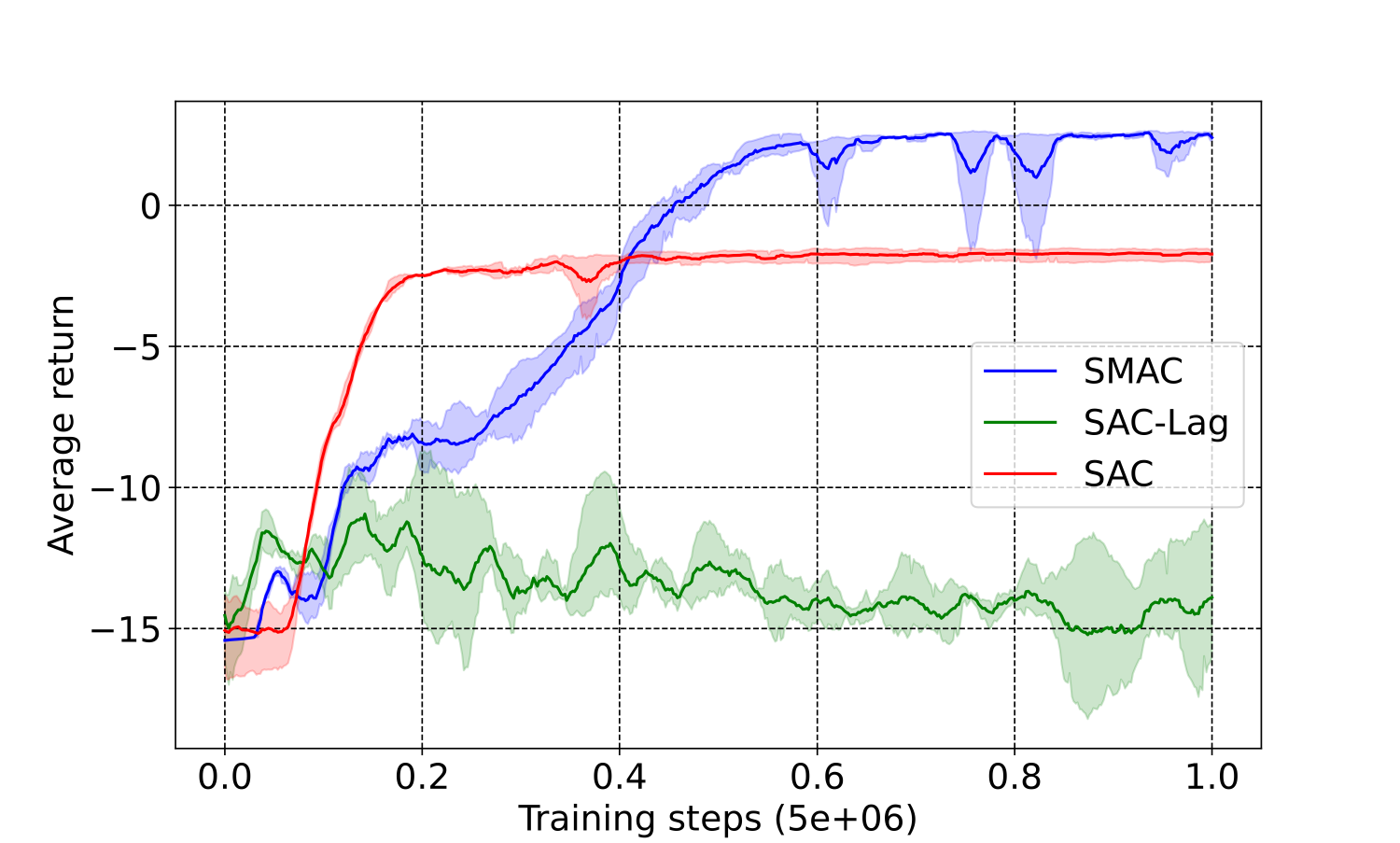}
\caption{The average return training curves of SAC, SAC-Lag, and SMAC by running 5 times. The lines and the shaded area represent the average return and the 95\% confidence interval, respectively.} \label{fig2}
\end{figure}

\begin{figure}[!h]
\centering
\includegraphics[scale=0.35]{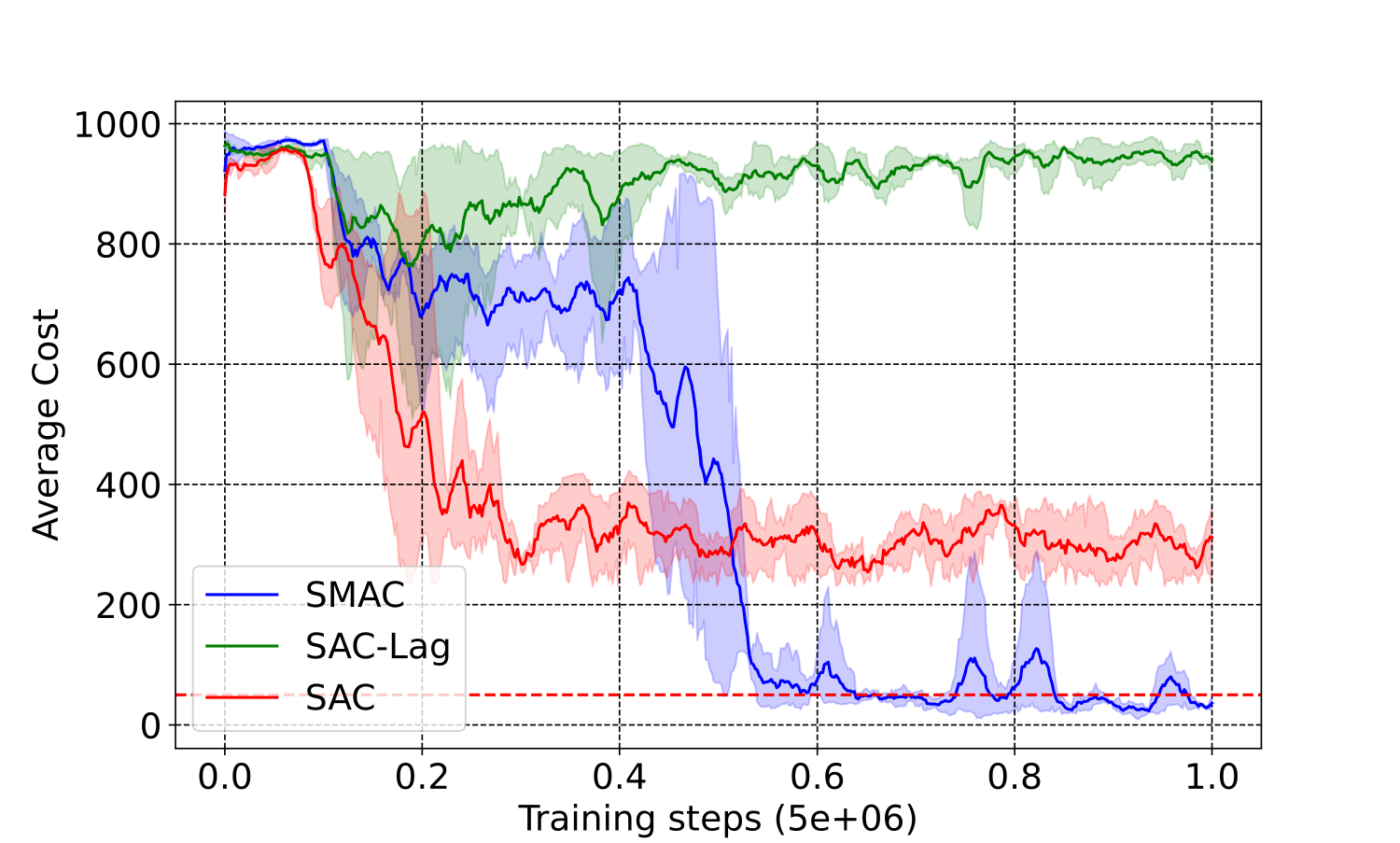}
\caption{The average cost training curves of SAC, SAC-Lag, and SMAC by running 5 times. The red dashed line is the safety constraint $C=50$.} \label{fig3}
\end{figure}

\begin{figure}[!h]
\centering
\includegraphics[scale=0.35]{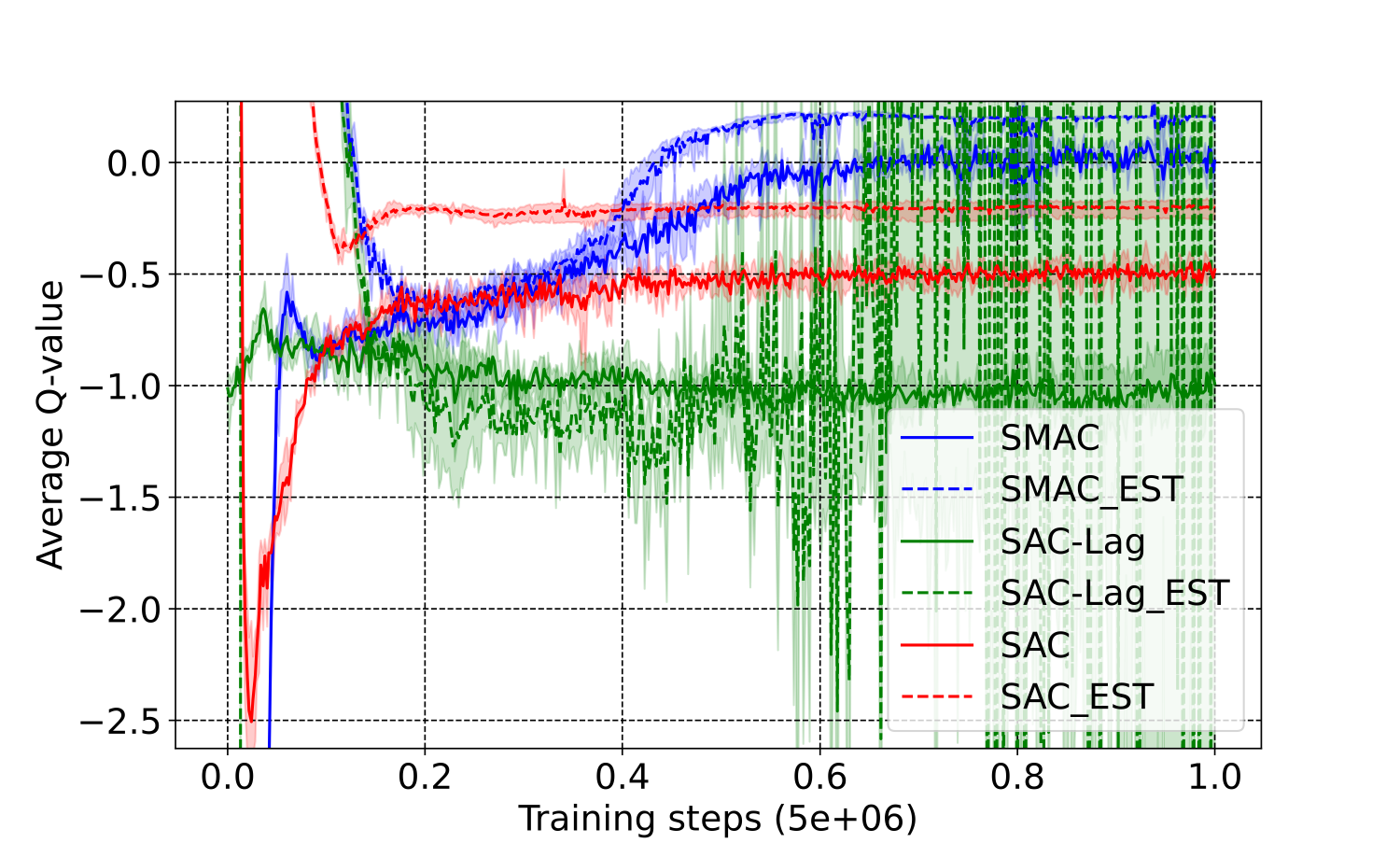}
\caption{The true average Q-value (solid lines) and estimated average Q-value (dashed lines) training curves by running 5 times at the 500th step per
episode.} \label{fig4}
\end{figure}

\begin{figure}[!h]
\centering
\includegraphics[scale=0.2]{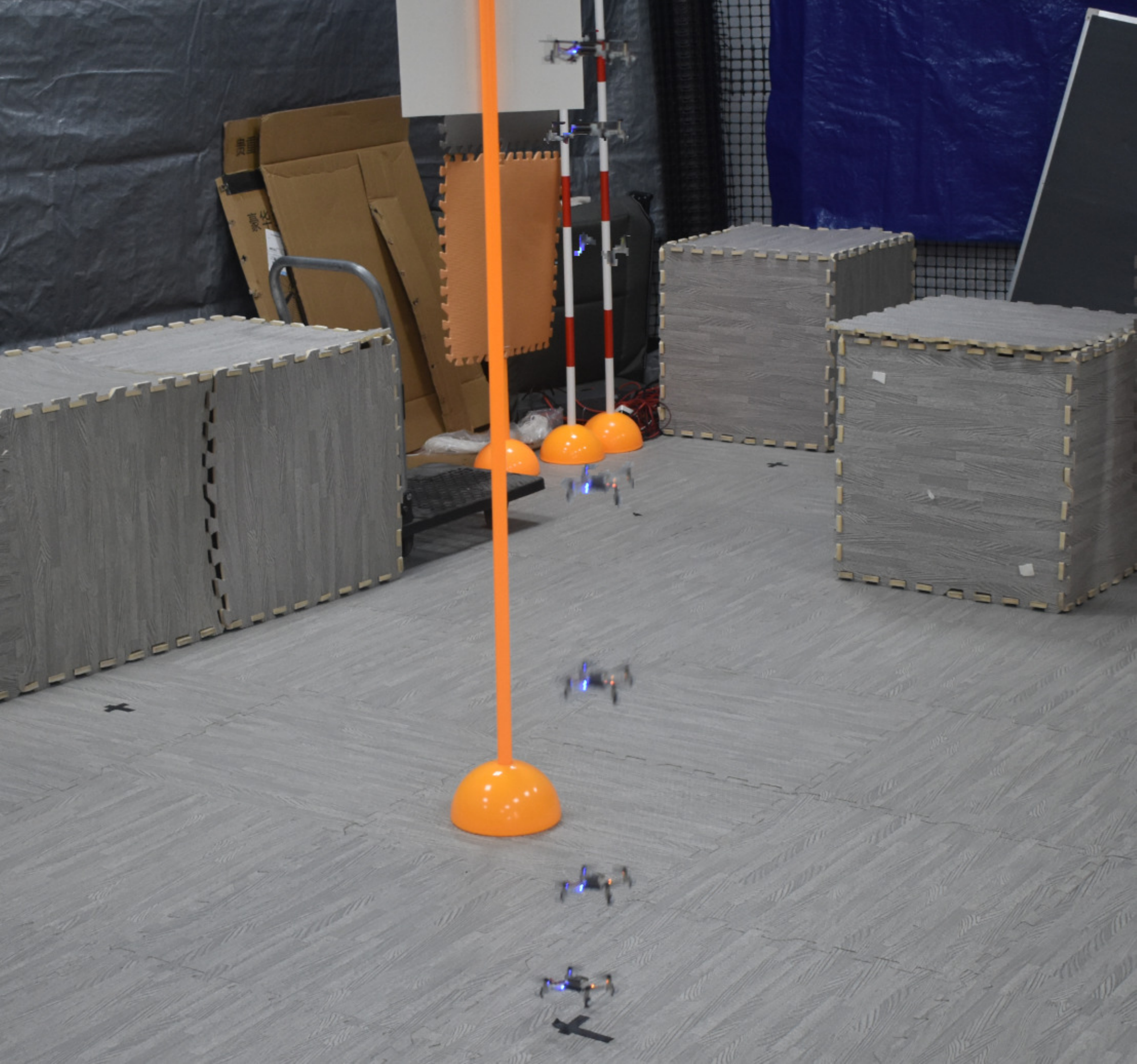}
\caption{The Crazyflie 2.1 hovering at 1.5m height in real-world.} \label{fig5}
\end{figure}

\subsubsection{Observation space}
The observation state $x_{t}$ is a 13-dimensional vector, which contains four parts: the distance between the target position and the current position $p_{t}=(p_{t}^{x}, p_{t}^{y}, p_{t}^{z})^{T}$, the current velocity $v_{t}=(v_{t}^{x}, v_{t}^{y}, v_{t}^{z})^{T}$, the current quaternion $R(\varrho_{t})$, where $\varrho_{t}=(\varrho_{t}^{r}, \varrho_{t}^{p}, \varrho_{t}^{\psi})^{T}$ is the current Euler angle, $R(\cdot)$ is the equation for converting Euler angle to quaternion, utilized to avoid gimbal lock, the Euler angular velocity $\omega_{t}=(\omega_{t}^{r}, \omega_{t}^{p}, \omega_{t}^{\psi})^{T}$.

\subsubsection{Action space}
The action $u_{t}\in[-u_{\max},u_{\max}]$ is a 4-dimensional vector, which is obtained from the modulation function $m(\bar{u}_{t}, \Delta u_{t})$, where $u_{\max}$ is the action bound. Inspired by \cite{yu2022towards, feng2024enhancing}, the corresponding actions and modulation function are designed as $\bar{u}_{t}=(a_{t},\varrho_{t}^{rc},\varrho_{t}^{pc},\varrho_{t}^{\psi c})^{T}$, where $a_{t}$ is the total acceleration command of the body's z-axis, $\varrho_{t}^{rc}$, $\varrho_{t}^{pc}$ and $\varrho_{t}^{\psi c}$ are the roll, pitch and yaw angle commands, respectively.  $\Delta u_{t}$ is the corresponding safety modulator for $\bar{u}_{t}$ and $m(\bar{u}_{t}, \Delta u_{t})=\bar{u}_{t}+\Delta u_{t}$.

\subsubsection{Reward \& cost reward design}
The reward function contains five parts: the distance reward $r^{dis}=-\|p_{t}\|$, the velocity reward $r^{vel}=-0.1\|v_{t}\|$, the action reward $r^{act}=-\|u_{t}\|$, the hit reward
$r^{hit}=\begin{cases}
-1, & \text{if hit the boundary},   \\
0, & \text{otherwise},
\end{cases}$ and the stay reward $r^{sta}=\begin{cases}
1.5, & \text{if} \ \|p_{t}\|<0.02, \\
0, & \text{otherwise}.
\end{cases}$

 The total reward function is defined as
\begin{align}\label{eq17}
r(x_{t},u_{t})=(r^{dis}+r^{vel}+r^{act}+r^{hit}+r^{sta})dt,
\end{align}
where $dt=1/240$ is the time step in PyBullet.

The cost reward function is designed to constrain Euler angles, which contains three parts: The roll angle cost reward $r_{c}^{r}$, the pitch angle cost reward $r_{c}^{p}$, and the yaw angle cost reward $r_{c}^{\psi}$, where\\
$r_{c}^{r}=\begin{cases}
1, & \text{if} \|\varrho_{t}^{r}\|<0.2, \\
0, & \text{otherwise},
\end{cases}$  $r_{c}^{p}=\begin{cases}
1, & \text{if} \|\varrho_{t}^{p}\|<0.2, \\
0, & \text{otherwise},
\end{cases}$  $r_{c}^{\psi}=\begin{cases}
1, & \text{if} \|\varrho_{t}^{\psi}\|<0.2, \\
0, & \text{otherwise}.
\end{cases}$

The total cost reward function is defined as
\begin{align}\label{eq18}
r_{c}(x_{t},u_{t})=r_{c}^{r}+r_{c}^{p}+r_{c}^{\psi}.
\end{align}

\begin{table*}
  \centering
  \caption{Training Parameters} \label{tab2}
\begin{tabular}{>{\centering\arraybackslash}m{0.2\textwidth} >{\centering\arraybackslash}m{0.2\textwidth} >{\centering\arraybackslash}m{0.2\textwidth}}
\toprule
Episode steps &  Training steps & Buffer size  \\
\hline
1000  & $5\times 10^{6}$ & $1\times 10^{6}$  \\
\hline
Target update $\tau$ & Discount factor $\gamma$ & Batch size \\
\hline
$5\times 10^{-3}$  &    0.99 &  512 \\
\hline
Safety constraint $C$ & Learning rate $\eta$ & Start learning step \\
\hline
$50$  &    $1\times 10^{-4}$ &  100 \\
\bottomrule
\end{tabular}
\end{table*}

\begin{figure}[!h]
  \centering
  \subfloat[SAC]{\includegraphics[width=0.5 \textwidth, trim=50 10 50 50,]{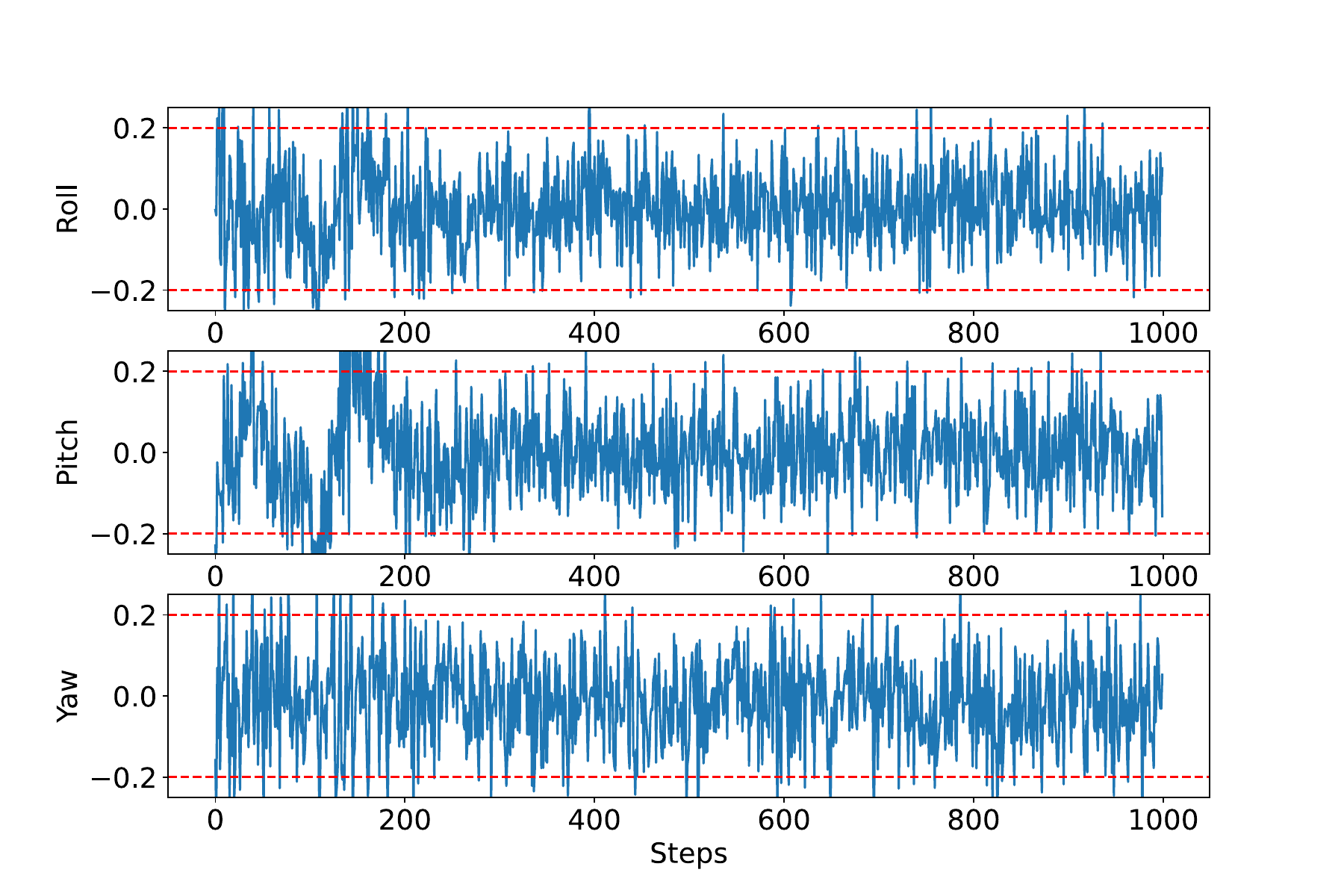}\label{fig:subfig7}}
   \vspace{0.1cm}
  \subfloat[SMAC]{\includegraphics[width=0.5 \textwidth, trim=50 25 50 10,]{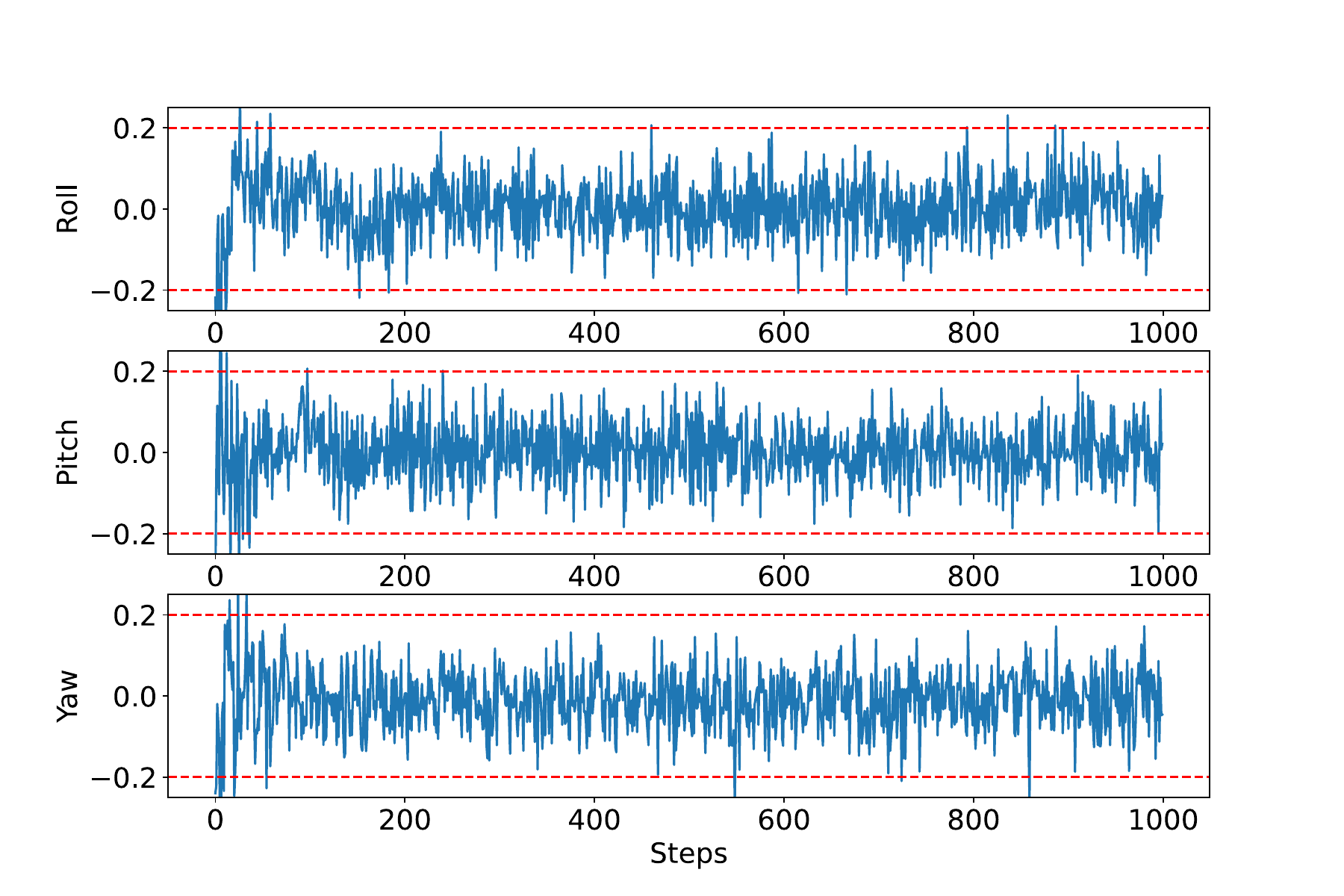}\label{fig:subfig8}}
  \caption{The roll, pitch, and yaw curves during hovering task using SAC (a) and SMAC (b).} \label{fig6}
\end{figure}
\subsection{Training results}
Before training, the risky policy, safety modulator, distributional critic, and cost critic networks are all modeled as 2-layer perceptrons with 256 hidden units. The activation function used in each unit is ReLU, and the final outputs of all networks are linear. The SMAC algorithm is designed on the Stable Baselines3. The training is conducted on a computer with an i7-13700K CPU and rendered with an RTX 4060ti GPU. The detailed training parameters are shown in \tabref{tab2}.

The simulation results of average return are shown in Fig. \ref{fig2}. Compared with other model-free methods SAC \cite{haarnoja2018soft} and SAC-Lag \cite{ha2020learning}, our SMAC makes the Crazyflie 2.1 hovering task achieve with a higher average return. The average cost results are shown in Fig. \ref{fig3}, where the safety constraint is $C=50$. Fig. \ref{fig2} and Fig. \ref{fig3} indicate that while SAC successfully achieves convergence of the return, it does not meet the safety constraint $C=50$. When SAC  takes the impact of safety constraints into account, it can be implemented through SAC-Lag. However, SAC-Lag leads to training failure
because the policy cannot trade off the maximization of rewards against cost rewards. Fortunately, with the help of safety modulator and distributional critic, the SMAC achieves the safety constraint and obtains the higher return than SAC.

To evaluate the effect of Q-value overestimation mitigation with distributional critic, we record the true Q-value and estimated Q-value by running five times with different seeds. Fig. \ref{fig4} shows the true Q-value and estimated Q-value curves during training. The Q-value is calculated once at the 500th steps per episode. Compared to SAC, SMAC exhibits a lower overestimation bias, indicating that the distributional critic effectively mitigates overestimation. Furthermore, SMAC achieves safety constraints with the help of safety modulator, whereas the consideration of safety constraints in SAC-Lag leads to divergent estimates Q-value because the policy fails to trade off the maximization of rewards against cost rewards, resulting in errors in the Q-value estimation.
\begin{table*}
  \centering
  \caption{The violation counts of roll, pitch, and yaw.} \label{tab3}
\begin{tabular}{>{\centering\arraybackslash}m{0.1\textwidth} >{\centering\arraybackslash}m{0.14\textwidth} >{\centering\arraybackslash}m{0.14\textwidth}  >{\centering\arraybackslash}m{0.14\textwidth} >{\centering\arraybackslash}m{0.14\textwidth} }
\toprule
Algorithms & roll  & pitch  & yaw & total  \\
\hline
SAC  & $72.80\pm8.87$ & $74.40\pm5.41$ & $95.00\pm4.18$ & $242.20\pm10.73$  \\
\hline
SMAC  & $20.40\pm2.70$ & $20.20\pm1.79$ & $7.20\pm2.59$ & $47.80\pm4.44$ \\
\bottomrule
\end{tabular}
\end{table*}

\subsection{Sim-to-Real}
With the help of precise simulation models of PyBullet, the hovering task can be easily deployed to the Crazyflie 2.1 in real-world as shown in Fig. \ref{fig5}. In real-world experiments, position and orientation information, such as Euler angles, are primarily calculated based on NOKOV Motion Capture System. As shown in Fig. \ref{fig6}, compared with SAC, the Crazyflie 2.1 controlled by the SMAC algorithm not only completes the task but also exhibits smaller fluctuations in the Euler angles, essentially meeting the safety constraints. After 5 rounds of testing, the  violation counts of the safety constraints by SMAC and SAC on the real-world device are presented in \tabref{tab3}. Regarding the total violation counts for roll, pitch, and yaw under safety constraints with $C=50$, with the help of the safety modulator, SMAC achieves safety constraints with a significantly smaller average total violation count of $47.80$. In contrast, SAC exhibits a substantially higher average count of $242.20$.

\section{Conclusions} \
In this paper, a new SMAC approach is proposed to address the issues of both safety and overestimation in model-free safe RL, where the safety modulator allows the policy to concentrate on maximizing rewards without the burden of trading off safety constraints. By introducing the theoretical update rule, the distributional critic can effectively mitigates overestimation. Both simulations and real-world scenarios demonstrate that the proposed SMAC strategy for UAV hovering task can maintain safety constraints and significantly outperforms existing baseline algorithms. This work paves the way for safer and more reliable deployment of model-free safe RL agents in real-world applications.

\bibliographystyle{ieeetran}   
\bibliography{rl}       

\begin{thebibliography}{10}
%
\bibitem{silverMasteringGameGo2016}
D.~Silver, A.~Huang, C.~J. Maddison, et al.,
``Mastering the game of {{Go}} with deep neural networks and tree search,''
\emph{Nature}, vol. 529, no. 7587, pp. 484--489, 2016.


\bibitem{sun2021model}
Y.~Sun, K.~Zhang, and C.~Sun, ``Model-based transfer reinforcement learning based on graphical model representations,'' \emph{IEEE Trans. Neural Netw. Learn. Syst.}, vol. 34, no. 2, pp. 1035-1048, 2023.


\bibitem{vinyalsGrandmasterLevelStarCraft2019}
O.~Vinyals, I.~Babuschkin, W.~M. Czarnecki,  et al., ``Grandmaster level in
{{StarCraft II}} using multi-agent reinforcement learning,'' \emph{Nature},
vol. 575, no. 7782, pp. 350--354, 2019.

\bibitem{tunyasuvunakool2020dmcontrol}
S.~Tunyasuvunakool, A.~Muldal, Y.~Doron,  et al., ``dm\_control: Software and tasks for
continuous control,'' \emph{Softw. Impacts}, vol.~6, p. 100022, 2020.

\bibitem{hao2023exploration}
J.~Hao, T.~Yang, H.~Tang, et al., ``Exploration in deep reinforcement learning: From single-agent to multiagent domain,'' \emph{IEEE Trans. Neural Netw. Learn. Syst.}, pp. 1-21, 2023.


\bibitem{gao2024reinforcement}
X.~Gao, J.~Si, and H.~Huang, ``Reinforcement learning control with knowledge shaping,'' \emph{IEEE Trans. Neural Netw. Learn. Syst.}, vol. 35, no. 3, pp. 3156-3167, 2024.



\bibitem{zhao2023state}
W.~Zhao, T.~He, R.~Chen,  et al., ``State-wise safe reinforcement
learning: A survey,'' \emph{arXiv preprint arXiv:2302.03122}, 2023.

\bibitem{dalal2018safe}
G.~Dalal, K.~Dvijotham, M.~Vecerik,  et al.,
``Safe exploration in continuous action spaces,'' \emph{arXiv preprint
arXiv:1801.08757}, 2018.

\bibitem{zhao2021model}
W.~Zhao, T.~He, and C.~Liu, ``Model-free safe control for zero-violation
reinforcement learning,'' in \emph{5th Annual Conference on Robot Learning},
2021.



\bibitem{tessler2018reward}
C.~Tessler, D.~J. Mankowitz, and S.~Mannor, ``Reward constrained policy
optimization,'' \emph{arXiv preprint arXiv:1805.11074}, 2018.

\bibitem{yang2023safety}
Q.~Yang, T.~D. Sim{\~a}o, S.~H. Tindemans,  et al.,
``Safety-constrained reinforcement learning with a distributional safety
critic,'' \emph{Mach. Learn.}, vol. 112, no.~3, pp. 859--887, 2023.

\bibitem{amir2024safe}
A.~Modares, N.~Sadati, B.~Esmaeili, et al., ``Safe reinforcement learning via a model-free safety certifier,'' \emph{IEEE Trans. Neural Netw. Learn. Syst.}, vol. 35, no. 3, pp. 3302--3311, 2024.


\bibitem{ma2024learn}
H.~Ma, C.~Liu, S.~E.~Li, et al., ``Learn zero-constraint-violation safe policy in model-free constrained reinforcement learning,'' \emph{IEEE Trans. Neural Netw. Learn. Syst.},  pp. 1-15, 2024.


\bibitem{wang2023safety}
H.~Wang, J.~Qin, and Z.~Kan, ``Shielded planning guided data-efficient and safe reinforcement learning,'' \emph{IEEE Trans. Neural Netw. Learn. Syst.}, pp. 1-12, 2024.


\bibitem{van2016deep}
H.~Van~Hasselt, A.~Guez, and D.~Silver, ``Deep reinforcement learning with
double Q-learning,'' in \emph{Proceedings of the AAAI Conference on
Artificial Intelligence}, vol.~30, no.~1, 2016.

\bibitem{lee2019bias}
D.~Lee and W.~B. Powell, ``Bias-corrected Q-learning with multistate
extension,'' \emph{IEEE Trans. Autom. Control}, vol.~64, no.~10,
pp. 4011--4023, 2019.

\bibitem{duan2021distributional}
J.~Duan, Y.~Guan, S.~E. Li,  et al., ``Distributional soft
actor-critic: Off-policy reinforcement learning for addressing value
estimation errors,'' \emph{IEEE Trans. Neural Netw. Learn. Syst.}, vol.~33, no.~11, pp. 6584--6598, 2021.

\bibitem{hasselt2010double}
H.~Hasselt, ``Double Q-learning,'' \emph{Advances in Neural Information
Processing Systems}, vol.~23, 2010.

\bibitem{fujimoto2018addressing}
S.~Fujimoto, H.~Hoof, and D.~Meger, ``Addressing function approximation error
in actor-critic methods,'' in \emph{International Conference on Machine
Learning}.\hskip 1em plus 0.5em minus 0.4em\relax PMLR, 2018, pp. 1587--1596.

\bibitem{du2023safe}
B.~Du, W.~Xie, Y.~Li,  et al.,
``Safe adaptive policy transfer reinforcement learning for distributed
multiagent control,'' \emph{IEEE Trans. Neural Netw. Learn. Syst.}, 2023.

\bibitem{haarnoja2018soft}
T.~Haarnoja, A.~Zhou, K.~Hartikainen,  et al., ``Soft actor-critic algorithms and
applications,'' \emph{arXiv preprint arXiv:1812.05905}, 2018.

\bibitem{ha2020learning}
S.~Ha, P.~Xu, Z.~Tan,  et al., ``Learning to walk in the real
world with minimal human effort,'' \emph{arXiv preprint arXiv:2002.08550},
2020.

\bibitem{altman2021constrained}
E.~Altman, \emph{Constrained Markov decision processes}.\hskip 1em plus 0.5em
minus 0.4em\relax Routledge, 2021.

\bibitem{wachi2020safe}
A.~Wachi and Y.~Sui, ``Safe reinforcement learning in constrained Markov
decision processes,'' in \emph{International Conference on Machine
Learning}.\hskip 1em plus 0.5em minus 0.4em\relax PMLR, 2020, pp. 9797--9806.

\bibitem{munos2016safe}
R.~Munos, T.~Stepleton, A.~Harutyunyan,  et al., ``Safe and efficient
off-policy reinforcement learning,'' \emph{Advances in Neural Information
Processing Systems}, vol.~29, 2016.

\bibitem{thrun2014issues}
S.~Thrun and A.~Schwartz, ``Issues in using function approximation for
reinforcement learning,'' in \emph{Proceedings of the 1993 connectionist
models summer school}.\hskip 1em plus 0.5em minus 0.4em\relax Psychology
Press, 2014, pp. 255--263.

\bibitem{bellemare2017distributional}
M.~G. Bellemare, W.~Dabney, and R.~Munos, ``A distributional perspective on
reinforcement learning,'' in \emph{International Conference on Machine
Learning}.\hskip 1em plus 0.5em minus 0.4em\relax PMLR, 2017, pp. 449--458.

\bibitem{barth2018distributed}
G.~Barth-Maron, M.~W. Hoffman, D.~Budden,  et al., ``Distributed distributional
deterministic policy gradients,'' \emph{arXiv preprint arXiv:1804.08617},
2018.

\bibitem{duan2023dsac}
J.~Duan, W.~Wang, L.~Xiao,  et al., ``Dsac-t: Distributional soft
actor-critic with three refinements,'' \emph{arXiv preprint
arXiv:2310.05858}, 2023.

\bibitem{yu2022towards}
H.~Yu, W.~Xu, and H.~Zhang, ``Towards safe reinforcement learning with a safety
editor policy,'' \emph{Advances in Neural Information Processing Systems},
vol.~35, pp. 2608--2621, 2022.

\bibitem{feng2024enhancing}
Y.~Feng, T.~Yang, and Y.~Yu, ``Enhancing UAV aerial docking: A hybrid approach
combining offline and online reinforcement learning,'' \emph{Drones}, vol.~8,
no.~5, p. 168, 2024.

\end{thebibliography}


\end{document}